%% file: acl_latex.tex
\pdfoutput=1
\documentclass[11pt]{article}
\usepackage[preprint]{acl}
\usepackage{times}
\usepackage{latexsym}
\usepackage[T1]{fontenc}
\usepackage[utf8]{inputenc}
\usepackage{microtype}
\usepackage{inconsolata}
\usepackage{xspace}
\usepackage{graphicx}
\usepackage{booktabs}
\usepackage{pifont}

\usepackage{amsmath}
\usepackage{tikz}
\usetikzlibrary{arrows.meta,positioning,calc,backgrounds,fit}

\DeclareRobustCommand{\method}{\textup{\textsc{TimeRSI}}\xspace}
\DeclareRobustCommand{\timebench}{\textit{TIME}\xspace}
\DeclareRobustCommand{\fevbench}{\textit{FEV-bench}\xspace}
\definecolor{scoregreen}{HTML}{21834D}
\definecolor{scorered}{HTML}{B43838}
\newcommand{\scorebetter}{\textcolor{scoregreen}{\ding{51}}}
\newcommand{\scoreworse}{\textcolor{scorered}{\ding{55}}}
\title{\method: Towards Self-Improved Agentic Deployment\\ for Time-Series Foundation Models}
\author{
  Haoxin Liu$^{1}$\thanks{\raggedright Contact us at \href{mailto:hliu763@gatech.edu}{\texttt{hliu763@gatech.edu}}, \href{mailto:badityap@cc.gatech.edu}{\texttt{badityap@cc.gatech.edu}}.} \quad
  Deqing Fu$^{2}$ \quad
  Lucheng Fu$^{1}$ \quad
  Yiqiao Jin$^{1}$ \\
  \textbf{Rushi Qiang$^{1}$} \quad
  \textbf{Harshavardhan Kamarthi$^{1,3}$} \quad
  \textbf{B. Aditya Prakash$^{1,3}$}\textsuperscript{\textasteriskcentered} \\
  $^{1}$Georgia Institute of Technology \quad
  $^{2}$University of Southern California \quad
  $^{3}$Samay AI
}

\begin{document}
\maketitle

\begin{abstract}
Time-series foundation models provide strong zero-shot forecast, but deployment for real-world applications still requires task-specific designs for model inputs and outputs. Direct inference can leave domain knowledge and recurring forecast errors unaddressed, while tailored pipelines require substantial human expertise and effort. We introduce \method, a self-improving agentic deployment system around frozen forecasting models. A coding agent translates task descriptions into executable pipeline changes. It tests each revision on historical forecasting windows and uses observed errors to guide subsequent revisions. During deployment, an adaptive selector chooses among the resulting pipelines using completed forecasts under similar conditions. Across 198 tasks on \timebench and \fevbench, \method outperforms all submissions on six point and probabilistic forecasting metrics. Controlled comparisons support the value of feedback-guided revision over non-iterative pipeline search. Further experiments verify the generalization with different coding agents and forecasting models. Code is available at \url{https://github.com/haoxin1998/TimeRSI}.
\end{abstract}

\input{figures/overview-float}

\section{Introduction}
Time-series foundation models (TSFMs) provide strong forecasts on new datasets without task-specific weight updates \citep{pmlr-v235-das24c,ansari2024chronoslearninglanguagetime,pmlr-v235-woo24a}. In real-world time-series forecasting (TSF), effective deployment still depends on how task knowledge shapes the model's inputs and outputs. For example, recurring holiday closures can inform demand forecasts, while known tidal cycles can guide ocean-current forecasts. Turning this knowledge into a \emph{deployment pipeline} around the model still requires substantial human expertise and effort.

Code agents based on large language models (LLMs) offer a way to automate pipeline development through program generation and evaluation \citep{jiang2025aideaidrivenexplorationspace,fu2026tabfmautoselfevolvingpipelinestabular}. Two challenges arise in forecasting. First, a plausible change may not improve accuracy. Calendar information, for instance, can enter through model inputs or a separate predictor, and these choices can yield different results. Historical forecast errors help evaluate and refine such decisions. Second, a pipeline that performs well on average may be less suitable as forecasting conditions change. 

We introduce \textbf{\method}, which applies recursive self-improvement (RSI) to deployment pipelines around frozen TSFMs (Fig.~\ref{fig:overview}). Our central idea is to use task knowledge to guide what to try and forecasting feedback to guide how to revise and use each pipeline. The agent edits input transformations, features, and forecast corrections while keeping the foundation model's weights fixed. 

Specifically, \method combines two stages. \emph{Iterative pipeline improvement} evaluates code revisions on historical forecasting windows and returns errors to the agent for further revision. A separate set of later historical windows identifies candidates for deployment. \emph{Adaptive pipeline selection} then chooses among these candidates for each new forecast using their performance under similar historical conditions.

We evaluate \method on \timebench and \fevbench \citep{qiao2026itstimegenerationtime,shchur2026fevbenchrealisticbenchmarktime}, covering \textbf{198 tasks}. \method \textbf{achieves the best scores among all existing submissions across six point and probabilistic forecasting metrics}. Ablations show that our iterative deployment improvement outperforms non-iterative pipeline search with the same number of candidates. Further experiments show gains with different coding agents and TSFMs, while case studies illustrate how task knowledge becomes effective deployment code.

\paragraph{Related work.}
Beyond direct deployment of foundation time-series models \citep{NEURIPS2024_662dcc5c,pmlr-v235-das24c,pmlr-v235-woo24a,ansari2024chronoslearninglanguagetime}, we study agentic deployment for realistic applications via recursive self-improvement (RSI) \citep{lee2026metaharnessendtoendoptimizationmodel,srikanth2026recursiveselfimprovementairesearch}. See App.~\ref{app:related} for additional related work.

\section{\method: Self-Improved Deployment for Time-Series Foundation Models}
\label{sec:method}

\method automates two deployment decisions: how to improve a task-specific pipeline and which pipeline to use for a new forecast. A code agent develops pipelines through repeated execution, evaluation, and revision. An adaptive selector chooses among them using past outcomes under similar forecasting conditions. The TSFM's weights remain fixed in both stages.

\input{figures/benchmark-float}

Formally, let $x_t$ contain the observed history and known covariates at forecast origin $t$, and let $m$ describe the task, variable meanings, and sampling frequency. A pipeline $p$ uses a frozen model $F_\theta$ to forecast the next $H$ steps as $\widehat{y}_{t+1:t+H} = p(F_\theta,x_t;m)$. The agent improves $p$, while the backbone parameters $\theta$ remain fixed.

The agent writes Python code across four stages: preprocessing, feature construction, context selection, and post-processing. On the input side, it can transform observations, encode task knowledge as additional variables, and choose which histories or variable groups the TSFM receives. On the output side, it can combine forecasts, fit auxiliary predictors, and correct systematic errors. This action space lets the agent translate task information into executable changes to the model's inputs and outputs. App.~\ref{app:interface} gives the programming interface.

\paragraph{Iterative Pipeline Improvement}
\label{sec:adaptation}

The outcome of a new forecast is not yet known, so \method uses completed historical windows to evaluate pipeline changes. For each window, the system supplies the history and known covariates at its forecast origin, runs the candidate, and scores its predictions for the following $H$ steps. Earlier windows form an adaptation set for iterative code revision; later windows form a selection set for choosing the deployment candidates.

The loop starts from direct TSFM inference as the reference pipeline. In each iteration, the agent uses the task description and earlier results to propose and implement a code revision. The system executes it on the adaptation windows and returns forecasting scores, window-level errors, and execution logs. The agent inspects this feedback to decide whether to refine its approach, combine useful changes, or try a different one. All earlier code and evaluation records remain accessible, so later revisions can build on successful ideas and address errors observed in previous attempts. Task semantics guide what to try, while execution feedback guides what to change next.

After the search budget is exhausted, the system evaluates the resulting candidates on the later selection windows. It retains the highest-ranked pipelines together with the reference to form a deployment set $\mathcal{P}$. This set provides several task-specific strategies for subsequent forecasts.

\paragraph{Adaptive Pipeline Selection}
\label{sec:deployment}

A pipeline that performs well on average may not be the best choice for every future window. \method therefore selects a pipeline for each target variable at each new forecast origin. The selector describes the current conditions using numerical features of the recent history and reference forecast, including level changes, recent trends, and forecast shape. It retrieves similar completed windows and compares how the retained pipelines performed on them.

For a given target, let $\mathcal{N}_t$ denote the retrieved windows whose outcomes are fully observed by $t$. Each window $s$ receives a similarity-based weight $w_{t,s}$. Let $\ell(p,s)$ be the error of pipeline $p$ on window $s$ relative to the reference. The selector chooses the pipeline with the lowest weighted historical error, $p_t^*=\arg\min_{p\in\mathcal{P}}\frac{\sum_{s\in\mathcal{N}_t}w_{t,s}\ell(p,s)}{\sum_{s\in\mathcal{N}_t}w_{t,s}}$.
The selected pipeline supplies the full forecast horizon and all quantiles for that target. App.~\ref{app:selector} specifies the features, weights, and loss.

At each origin, \method runs and records forecasts from every retained pipeline. Once a window is complete, the system adds each pipeline's errors to its feedback history and uses them to inform subsequent choices. Pipeline code remains fixed during this stage, while selection adapts as new outcomes arrive. Thus, iterative revision improves the available strategies, and adaptive selection determines when to use them.

\section{Experiments}
\label{sec:experiments}
\input{figures/cases-and-ablation}
\input{sections/benchmark-experiments}

\textbf{Generalization. Effective across coding agents and TSFMs.}
The gains extend beyond our default components (Tab.~\ref{tab:generalization}). With TimesFM~3.0 fixed, Antigravity (Gemini~3.8~Flash) reaches normalized MASE/CRPS of 0.637/0.532, outperforming the compared external submissions on \timebench. With Claude~Code fixed, \method reduces MASE/CRPS from 0.662/0.556 to 0.647/0.545 for Chronos-2 and from 0.669/0.572 to 0.641/0.549 for LPTM, whose \fevbench MASE/SQL Skill also rises from 30.70/43.20 to 34.63/46.33 (Fig.~\ref{fig:benchmarks}). The Claude~Code--TimesFM~3.0 combination achieves better results, consistent with the advantages of more advanced coding agents and TSFMs.

\paragraph{Case studies.}
The agent translates task knowledge into pipeline changes (Fig.~\ref{fig:deployment-cases}): calendar features for parking, tidal decomposition for ocean currents, history-supported holiday corrections for demand, and daily-reset shape corrections for solar counters. The resulting pipelines reduce task-level MASE by 2.86--6.70\%. The parking case also illustrates feedback-guided revision: after a calendar-predictor blend increases historical error, the agent switches to calendar inputs and adds weekend--time interactions. Feedback thus guides changes in strategy, not just parameter tuning. App.~\ref{app:cases} provides full traces and additional examples.
\section{Conclusion}
We introduce \method to automate task-specific deployment through iterative pipeline improvement and adaptive selection around frozen TSFMs. On \timebench and \fevbench, \method ranks first across six forecasting metrics compared with competitive external submissions.
\section*{Limitations}
 \method relies on clear task and variable descriptions to connect numerical patterns with domain knowledge. Besides, gains on historical windows do not guarantee gains on future forecasts. This highlights the open problem, i.e., distribution shift, in time-series forecasting \citep{pmlr-v235-liu24ae,Kim2025}.

\section*{Acknowledgements}
This paper was supported in part by the NSF (Expeditions CCF-1918770, CAREER IIS-2028586, Medium IIS-1955883, Medium IIS-2403240, Medium IIS-2106961), NIH (1R01HL184139), CDC MInD program, Meta, and Dolby faculty gifts.

\input{references}
\appendix
\clearpage
\section{Additional Related Work}
\label{app:related}

\paragraph{Pretrained forecasting models.}
TimesFM \citep{pmlr-v235-das24c}, Chronos \citep{ansari2024chronoslearninglanguagetime}, and Moirai \citep{pmlr-v235-woo24a} learn from diverse numerical series to support forecasting on new datasets. LPTM \citep{NEURIPS2024_662dcc5c} studies adaptive segmentation during pretraining for cross-domain time-series analysis. These works develop the numerical backbone. \method addresses a complementary question: how can an agent adapt the deployment pipeline around an existing backbone while keeping its weights fixed?

\paragraph{Semantics and temporal generalization.}
LSTPrompt \citep{liu-etal-2024-lstprompt} studies structured prompting for time-series forecasting. VL-Time \citep{liu-etal-2025-picture} uses visual representations to support language-guided time-series reasoning. Time-MMD \citep{NEURIPS2024_8e776812} and TimesX \citep{pmlr-v306-liu26go} provide benchmarks connecting numerical series with textual context. These studies motivate using information beyond raw target values; our agent turns task information into executable pipeline changes. FOIL \citep{pmlr-v235-liu24ae} addresses out-of-distribution forecasting through invariant learning. Rather than learning invariant backbone representations, \method uses completed forecasts to choose among candidate pipelines.

\paragraph{Agent improvement.}
Self-Refine \citep{NEURIPS2023_91edff07} alternates generation, self-feedback, and revision. Reflexion \citep{NEURIPS2023_1b44b878} carries verbal reflections from task feedback into later attempts. GEPA \citep{ICLR2026_0e9e708b} evolves prompts using execution traces and reflective feedback. Adaptive Auto-Harness \citep{liu2026adaptiveautoharnesssustainedselfimprovement} combines harness evolution and task-wise routing on open-ended streams. \method focuses on executable pipelines around a frozen numerical forecaster. It separates historical code search from deployment selection, where completed forecast windows supply the new feedback.

\paragraph{Machine learning engineering agents.}
MLAgentBench \citep{pmlr-v235-huang24y} evaluates agents that read and write files, execute experiments, and inspect outputs. MLE-bench \citep{ICLR2025_7e3767db} assesses model development and experimentation on machine learning competitions. AIDE \citep{jiang2025aideaidrivenexplorationspace} searches over code solutions, while TabFM-Auto \citep{fu2026tabfmautoselfevolvingpipelinestabular} focuses pipeline search around a frozen tabular model. Our design shares the code--execution--feedback pattern, but couples pipeline adaptation with chronological feedback and strategy selection for repeated forecasting.

\section{Pipeline Interface}
\label{app:interface}

Four functions pass prepared data and state between pipeline stages. The first, \texttt{preprocess}, receives the observed history and task description and returns prepared data and local state. \texttt{engineer} constructs variables from these inputs. \texttt{select\_context} creates one or more forecasting cases, specifying the histories and variable groups supplied to the foundation model. \texttt{postprocess} maps model outputs back to the target scale and returns point and quantile forecasts in the original target units. State is local to the current invocation. It can hold fitted transforms or auxiliary predictors derived from the available history. During deployment, these components may be fitted again on newly available history while their code remains fixed.

The evaluator supplies the same chronological windows to all candidates. Candidate code receives no target values beyond each window's origin. The agent can access the labels of the adaptation windows through evaluation feedback; these labels are not passed to the pipeline as prediction inputs. Later selection feedback is used by the system only after code search finishes. All four functions and their evaluation records are retained so the agent can inspect earlier attempts.

\section{Historical Selection Details}
\label{app:selector}

The reference configuration describes each target using up to 96 recent observations and the reference median forecast. Its six features are: the latest value relative to the recent median; the mean of the last eight observations relative to the mean of the earlier half of the recent history; the first forecast relative to the latest observation; the mean forecast relative to the recent median; the standard deviation across the forecast horizon; and the last forecast relative to the first. Each feature is scaled by the recent historical standard deviation, with a small floor for nearly constant series.

At a new origin, the selector first filters out windows whose outcomes are not yet complete. It uses records for the current item when at least three distinct origins are available; otherwise it pools the available records. For each target, it standardizes the feature coordinates using the historical feature standard deviations, floored at $0.25$. Distance is the mean squared difference in these coordinates. The selector retrieves up to 12 nearest records and gives each an initial weight of $1/(1+d)$, where $d$ is this distance. Weights are then normalized within each origin before applying the weighted-error selection rule.

For each candidate and historical window, the loss is its mean absolute scaled error (MASE) divided by the reference MASE, with a denominator floor of $10^{-8}$. MASE divides mean absolute forecast error by the mean absolute seasonal difference in the history available at that origin. Candidate ranking before deployment uses mean selection-window MASE, with the quantile score as a tie-breaker. The portable configuration retains three ranked candidates plus the reference if it is not already among them. Initial feedback comes from the completed adaptation and selection windows; subsequent completed forecasts extend this history. With no available feedback, the selector uses the reference pipeline.

\input{sections/discovered-pipelines}

\input{sections/benchmark-details}

\section{Use of AI Assistants}
\label{app:ai-assistants}

AI assistants were used to refine language, improve manuscript organization, and support coding and experimental workflows. The authors guided the research and take responsibility for the study design, methodological choices, reported results, and final manuscript.

\end{document}

%% file: figures/overview-float.tex
\begin{figure*}[!t]
  \centering
  \resizebox{\textwidth}{!}{\input{figures/overview}}
  \caption{\textbf{From conventional to self-improved TSFM deployment.}
  (a) Direct inference leaves task-specific adaptation to the user, while tailored pipelines require human design and revision.
  (b) \method automates pipeline improvement through task-aware code edits and historical forecasting feedback, then selects among the resulting pipelines for each new forecast. The TSFM remains frozen.}
  \label{fig:overview}

\end{figure*}

%% file: figures/overview.tex
\definecolor{ice}{HTML}{2775A4}
\definecolor{icefill}{HTML}{E8F4FC}
\definecolor{fire}{HTML}{B95728}
\definecolor{firefill}{HTML}{FFF0E2}
\definecolor{worldfill}{HTML}{F3F6F8}
\definecolor{ink}{HTML}{263A48}
\definecolor{muted}{HTML}{697B87}
\tikzset{
  pics/smiling/.style={code={
    \draw[black,line width=0.65pt] (0,0) circle (0.15);
    \fill[black] (-0.052,0.035) circle (0.014);
    \fill[black] (0.052,0.035) circle (0.014);
    \draw[black,line width=0.6pt,line cap=round]
      (-0.065,-0.035) .. controls (0,-0.12) .. (0.065,-0.035);
  }},
  pics/crying/.style={code={
    \draw[black,line width=0.65pt] (0,0) circle (0.15);
    \fill[black] (-0.052,0.035) circle (0.014);
    \fill[black] (0.052,0.035) circle (0.014);
    \draw[black,line width=0.6pt,line cap=round]
      (-0.055,-0.075) .. controls (0,-0.02) .. (0.055,-0.075);
    \fill[ice] (0.09,0.005) .. controls (0.13,-0.055) and
      (0.05,-0.065) .. (0.09,0.005);
  }},
  pics/human/.style={code={
    \filldraw[fill=white,draw=muted,line width=0.9pt] (0,0.12) circle (0.105);
    \filldraw[fill=white,draw=muted,line width=0.9pt,rounded corners=1pt]
      (-0.22,-0.21) -- (-0.22,-0.12)
      .. controls (-0.22,0.01) and (0.22,0.01) .. (0.22,-0.12)
      -- (0.22,-0.21) -- cycle;
  }},
  pics/snowflake/.style={code={
    \foreach \a in {0,60,...,300}{
      \draw[ice,line width=0.85pt,line cap=round] (0,0) -- (\a:0.20);
      \draw[ice,line width=0.65pt,line cap=round]
        (\a:0.12) -- ++({\a+55}:0.075)
        (\a:0.12) -- ++({\a-55}:0.075);
    }
  }},
  pics/robot/.style={code={
    \draw[fire,line width=0.8pt] (0,0.19) -- (0,0.30);
    \fill[fire] (0,0.31) circle (0.035);
    \filldraw[fill=firefill,draw=fire,line width=0.8pt,rounded corners=1.5pt]
      (-0.23,-0.17) rectangle (0.23,0.20);
    \draw[fire,line width=1.7pt,line cap=round]
      (-0.28,-0.06) -- (-0.28,0.08)
      (0.28,-0.06) -- (0.28,0.08);
    \fill[ink] (-0.085,0.065) circle (0.035);
    \fill[ink] (0.085,0.065) circle (0.035);
    \draw[fire,line width=0.7pt,line cap=round] (-0.09,-0.075) -- (0.09,-0.075);
  }}
}
\begin{tikzpicture}[
  x=1cm,y=1cm,
  every node/.style={font=\fontsize{9}{10.5}\selectfont,text=ink,
    align=center,inner sep=2pt},
  head/.style={font=\fontsize{9.7}{11}\selectfont\bfseries,anchor=west},
  flow/.style={-{Latex[length=1.8mm]},draw=muted,line width=0.8pt},
  hotflow/.style={-{Latex[length=1.8mm]},draw=fire,line width=1pt},
  stage/.style={rounded corners=3pt,fill=white,draw=muted!40,
    minimum width=3.05cm,minimum height=0.56cm},
  process/.style={rounded corners=4pt,draw=fire!60,fill=white,
    minimum height=0.92cm}
]
\path[use as bounding box] (-0.02,0.10) rectangle (17.55,7.55);

\node[head,font=\fontsize{9.7}{11}\selectfont\bfseries] at (0,7.25)
  {(a) Conventional Deployment};
\node[head,text=fire] (oursheading) at (4.70,7.25)
  {(b) \method: Self-Improved Agentic Deployment};
\pic[scale=0.95] at ($(oursheading.east)+(0.25,0)$) {smiling};
\draw[muted!35,line width=0.8pt] (0,6.94) -- (4.25,6.94);
\draw[fire!65,line width=0.8pt] (4.70,6.94) -- (17.50,6.94);

\filldraw[fill=firefill!22,draw=fire!25,rounded corners=7pt,line width=0.7pt]
  (4.55,0.95) rectangle (17.50,6.60);

\filldraw[fill=worldfill,draw=muted!30,rounded corners=6pt]
  (0,4.00) rectangle (4.25,6.57);
\node[font=\fontsize{9}{10.5}\selectfont\bfseries] at (2.125,6.25)
  {Real-World TSF Applications};
\node[font=\fontsize{8.5}{10}\selectfont\bfseries] at (2.125,5.93)
  {Out-of-the-box inference};
\filldraw[fill=icefill,draw=ice,line width=1pt,rounded corners=4pt]
  (0.60,4.96) rectangle (3.65,5.62);
\pic[scale=0.70] at (1.02,5.29) {snowflake};
\node[text=ice,font=\fontsize{9.2}{10.6}\selectfont\bfseries] at (2.31,5.29)
  {TSFM};
\node[font=\fontsize{8.1}{9.5}\selectfont,text=black] at (1.96,4.47)
  {Limited task-specific\\adaptation};
\pic[scale=0.85] at (3.60,4.32) {crying};

\filldraw[fill=worldfill,draw=muted!30,rounded corners=6pt]
  (0,0.98) rectangle (4.25,3.67);
\node[font=\fontsize{9}{10.5}\selectfont\bfseries] at (2.125,3.35)
  {Real-World TSF Applications};
\filldraw[fill=white,draw=muted!55,rounded corners=5pt,line width=1pt]
  (0.40,1.59) rectangle (3.85,2.99);
\pic[scale=0.62] at (0.77,2.70) {human};
\node[font=\fontsize{8.3}{9}\selectfont\bfseries,text=muted]
  at (2.30,2.68) {Human-designed\\pipeline};
\filldraw[fill=icefill,draw=ice,line width=1pt,rounded corners=4pt]
  (0.60,1.80) rectangle (3.65,2.35);
\pic[scale=0.60] at (1.02,2.075) {snowflake};
\node[text=ice,font=\fontsize{8.8}{10}\selectfont\bfseries] at (2.31,2.075)
  {TSFM};
\node[font=\fontsize{8.1}{9.5}\selectfont,text=black] at (1.93,1.27)
  {High human effort};
\pic[scale=0.85] at (3.24,1.27) {crying};

\filldraw[fill=worldfill,draw=muted!25,rounded corners=5pt]
  (4.80,1.15) rectangle (9.05,6.40);
\node[font=\fontsize{9}{10.5}\selectfont\bfseries] at (6.925,6.09)
  {Real-World TSF Applications};
\filldraw[fill=firefill,draw=fire,rounded corners=6pt,line width=1.3pt]
  (5.04,1.92) rectangle (8.81,5.68);
\pic[scale=0.58] at (5.39,5.33) {robot};
\node[font=\fontsize{8.1}{9.5}\selectfont\bfseries,text=fire] at (7.22,5.33)
  {Agent-improved pipeline};
\node[stage,draw=fire!55] (inputs) at (6.925,4.68)
  {Features \& model inputs};
\filldraw[fill=icefill,draw=ice,line width=1pt,rounded corners=4pt]
  (5.40,3.06) rectangle (8.45,4.08);
\pic[scale=0.72] at (5.695,3.78) {snowflake};
\node[text=ice,font=\fontsize{9.3}{10.6}\selectfont\bfseries,
  text width=2.92cm] at (6.925,3.57)
  {Time-series\\foundation model};
\node[stage,draw=fire!55] (outputs) at (6.925,2.47) {Forecast corrections};
\draw[hotflow] (inputs.south) -- (6.925,4.12);
\draw[hotflow] (6.925,3.02) -- (outputs.north);
\draw[hotflow] (outputs.south) -- (6.925,1.72);
\node[text=fire] at (6.925,1.45) {Forecasts};

\fill[fire!5] (8.82,5.66) -- (9.65,6.37) -- (9.65,1.18)
  -- (8.82,1.94) -- cycle;
\draw[fire!55,dashed,line width=0.7pt] (8.82,5.66) -- (9.65,6.37);
\draw[fire!55,dashed,line width=0.7pt] (8.82,1.94) -- (9.65,1.18);

\filldraw[fill=white,draw=fire!45,rounded corners=5pt,line width=0.8pt]
  (9.65,3.78) rectangle (17.25,6.40);
\node[anchor=west,font=\fontsize{10}{11.5}\selectfont\bfseries,text=fire]
  at (9.88,6.06) {1\quad Iterative Pipeline Improvement};
\node[text=muted] (semantics) at (11.70,5.54) {Task semantics};
\node[process,minimum width=3.25cm,fill=firefill!65] (agent) at (11.70,4.79) {};
\pic[scale=0.66] at (10.45,4.90) {robot};
\node[font=\fontsize{9}{10.5}\selectfont\bfseries,text=fire] at (11.96,4.94)
  {Deployment Agent};
\node[font=\fontsize{8.7}{10}\selectfont] at (11.70,4.57)
  {Revise pipeline code};
\node[process,minimum width=2.97cm] (evaluate) at (15.48,4.79)
  {\textbf{Evaluate pipeline}\\on historical windows};
\draw[flow] (semantics.south) -- (agent.north);
\draw[hotflow] (agent.east) -- (evaluate.west);
\draw[hotflow] (evaluate.south) -- (15.48,4.10) -- (11.70,4.10) -- (agent.south);
\node[fill=white,text=fire,font=\fontsize{8.7}{10}\selectfont] at (13.59,4.10)
  {Error feedback};

\filldraw[fill=white,draw=fire!45,rounded corners=5pt,line width=0.8pt]
  (9.65,1.15) rectangle (17.25,3.12);
\node[anchor=west,font=\fontsize{10}{11.5}\selectfont\bfseries,text=fire]
  at (9.88,2.82) {2\quad Adaptive Pipeline Selection};
\node[font=\fontsize{8.3}{10}\selectfont,text width=3.15cm] (scenarios) at (11.30,1.98)
  {Forecast conditions\\+ historical feedback};
\node[process,minimum width=1.25cm,minimum height=0.64cm,fill=firefill!65]
  (select) at (14.22,1.98) {\textbf{Select}};
\node[font=\fontsize{9}{10.5}\selectfont\bfseries] (forecast) at (16.20,1.98)
  {Forecast};
\draw[flow] (scenarios.east) -- (select.west);
\draw[hotflow] (select.east) -- (forecast.west);

\draw[hotflow] (16.85,3.78) -- (16.85,2.45) -- (14.22,2.45) -- (select.north);
\node[text=fire,font=\fontsize{8.7}{10}\selectfont] at (14.55,3.44)
  {Candidate pipelines};

\pic[scale=0.72] at (4.22,0.32) {snowflake};
\node[anchor=west,text=ice,font=\fontsize{8.7}{10}\selectfont] at (4.49,0.32)
  {Frozen model weights};
\pic[scale=0.60] at (9.29,0.32) {robot};
\node[anchor=west,text=fire,font=\fontsize{8.7}{10}\selectfont] at (9.59,0.32)
  {Agent-edited pipeline code};
\end{tikzpicture}

%% file: figures/benchmark-float.tex
\begin{figure*}[!t]
  \centering
  \includegraphics[width=\textwidth]{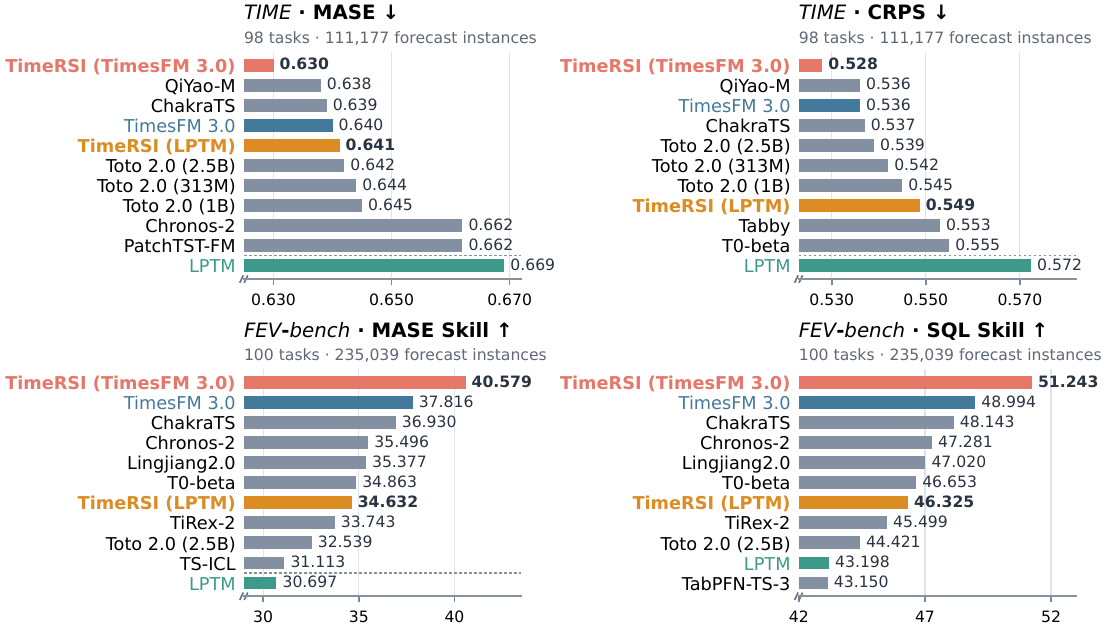}
  \caption{\textbf{\method ranks first among the compared submissions across six reported metrics on \timebench and \fevbench.} Labels name the frozen backbone of \method; LPTM~\citep{NEURIPS2024_662dcc5c} is appended below the dashed line when outside the top ten. See App.~\ref{app:benchmark-details} for metrics including WQL Skill and WAPE Skill, and full results of more methods.}
  \label{fig:benchmarks}
\end{figure*}

%% file: figures/cases-and-ablation.tex
\begin{figure*}[!t]
  \centering
  \input{figures/deployment-cases}
  \caption{\textbf{Agentic deployment examples via \method.} The agent adds calendar features for parking, separates tidal patterns for ocean currents, applies holiday rules for supply-chain demand, and restores daily accumulation and reset patterns for solar counters. These agentic deployments reduce MASE by 2.86--6.70\% over direct TimesFM~3.0.}
  \label{fig:deployment-cases}

  \par\vspace{8pt}
  \begingroup
  \captionsetup{type=table}
  \small
  \resizebox{\textwidth}{!}{\begin{tabular*}{\textwidth}{@{\extracolsep{\fill}}lllll@{}}
    \toprule
    & \multicolumn{2}{c}{\textbf{\timebench}: normalized errors}
    & \multicolumn{2}{c}{\textbf{\fevbench}: skill scores (\%)} \\
    \cmidrule(lr){2-3}\cmidrule(lr){4-5}
    Method & MASE $\downarrow$ & CRPS $\downarrow$
    & SQL Skill $\uparrow$ & MASE Skill $\uparrow$ \\
    \midrule
    Direct TimesFM~3.0 & 0.640 & 0.536 & 48.984 & 37.786 \\
    Non-Iterative Search for Deployment & 0.650\,\scoreworse & 0.549\,\scoreworse
    & 42.832\,\scoreworse & 30.509\,\scoreworse \\
    \method Deployment & \textbf{0.630}\,\scorebetter & \textbf{0.528}\,\scorebetter
    & \textbf{51.243}\,\scorebetter & \textbf{40.579}\,\scorebetter \\
    \bottomrule
  \end{tabular*}
  }
  \caption{\textbf{Ablation of iterative deployment improvement.} With the same number of candidate pipelines, non-iterative search without feedback-guided revision underperforms direct TimesFM~3.0 inference, while \method improves performance on both benchmarks.}
  \label{tab:feedback-ablation}
  \endgroup
\end{figure*}

%% file: figures/deployment-cases.tex
\definecolor{caseink}{HTML}{243748}
\definecolor{casemuted}{HTML}{5E6B76}
\definecolor{caseline}{HTML}{D9E1E7}
\definecolor{caseblue}{HTML}{2775A4}
\definecolor{caseteal}{HTML}{147D78}
\definecolor{caseamber}{HTML}{AA5D22}
\definecolor{caseviolet}{HTML}{70539A}
\definecolor{casegain}{HTML}{166B47}
\definecolor{casegainfill}{HTML}{ECF6F0}
\begin{tikzpicture}[
  x=1cm,y=1cm,
  every node/.style={text=caseink,inner sep=0pt,outer sep=0pt},
  casetitle/.style={font=\fontsize{8.8}{10}\selectfont\bfseries,anchor=west},
  casekind/.style={font=\fontsize{8}{9.5}\selectfont,anchor=west},
  casecue/.style={font=\fontsize{8.2}{10}\selectfont,anchor=north west,
    align=left,text width=3.39cm},
  casecode/.style={font=\fontsize{7.2}{9.6}\selectfont\ttfamily,
    anchor=north west,align=left},
  caseflow/.style={font=\fontsize{8}{9.5}\selectfont,align=center},
  casescore/.style={font=\fontsize{8.2}{9.5}\selectfont},
  casegaintext/.style={font=\fontsize{9}{10}\selectfont\bfseries,text=casegain}
]
\path[use as bounding box] (0,0) rectangle (15.96,5.27);

\node[font=\fontsize{9}{10.5}\selectfont\bfseries,anchor=west,text=caseblue]
  at (0.04,5.06) {INPUT-SIDE ADAPTATION};
\node[font=\fontsize{9}{10.5}\selectfont\bfseries,anchor=west,text=caseamber]
  at (8.14,5.06) {OUTPUT-SIDE ADAPTATION};
\draw[caseblue!35,line width=0.65pt] (0.04,4.87) -- (7.80,4.87);
\draw[caseamber!35,line width=0.65pt] (8.14,4.87) -- (15.92,4.87);

\foreach \x/\accent in {0/caseteal,4.05/caseblue,8.10/caseamber,12.15/caseviolet}{
  \begin{scope}[shift={(\x,0)}]
    \filldraw[fill=white,draw=caseline,line width=0.65pt,rounded corners=3pt]
      (0,0.02) rectangle (3.81,4.73);
    \fill[\accent,rounded corners=1pt] (0.01,4.65) rectangle (3.80,4.73);
    \fill[\accent!7,rounded corners=2pt] (0.15,2.48) rectangle (3.66,3.07);
    \fill[casegainfill,rounded corners=2pt] (0.15,0.16) rectangle (3.66,0.89);
  \end{scope}
}

\begin{scope}
  \node[casetitle,text=caseteal] at (0.20,4.43) {(a) Parking availability};
  \node[casekind,text=caseteal] at (0.20,4.10) {Task-specific features};
  \node[casecue] at (0.20,3.78)
    {Weekday and weekend\\parking patterns differ.};
  \node[caseflow] at (1.905,2.775)
    {Add calendar features\\$\downarrow$ Frozen TSFM};
  \node[casecode] at (0.20,2.20)
    {x = calendar(dates)\\
     x += weekend\_time(dates)\\
     q = {\color{caseblue}TimesFM}(y, x)};
  \node[casescore] at (1.905,0.65) {MASE: 0.679 $\rightarrow$ \textbf{0.642}};
  \node[casegaintext] at (1.905,0.33) {$\downarrow$\,5.46\%};
\end{scope}

\begin{scope}[shift={(4.05,0)}]
  \node[casetitle,text=caseblue] at (0.20,4.43) {(b) Ocean currents};
  \node[casekind,text=caseblue] at (0.20,4.10) {Domain knowledge};
  \node[casecue] at (0.20,3.78)
    {Tidal cycles explain\\part of current variation.};
  \node[caseflow] at (1.905,2.775)
    {Remove tide $\rightarrow$ TSFM\\$\rightarrow$ Add tide};
  \node[casecode] at (0.20,2.20)
    {tide = fit\_tides(y)\\
     r = {\color{caseblue}TimesFM}(y - tide.past)\\
     q = r + tide.future};
  \node[casescore] at (1.905,0.65) {MASE: 0.872 $\rightarrow$ \textbf{0.835}};
  \node[casegaintext] at (1.905,0.33) {$\downarrow$\,4.18\%};
\end{scope}

\begin{scope}[shift={(8.10,0)}]
  \node[casetitle,text=caseamber] at (0.20,4.43) {(c) Supply-chain demand};
  \node[casekind,text=caseamber] at (0.20,4.10) {Business rules};
  \node[casecue] at (0.20,3.78)
    {Certain holidays show\\repeated zero demand.};
  \node[caseflow] at (1.905,2.775)
    {Forecast $\rightarrow$ Apply\\history-supported rule};
  \node[casecode] at (0.20,2.20)
    {days = checked\_holidays(y)\\
     q = candidate\_forecast\\
     q[days] = 0};
  \node[casescore] at (1.905,0.65) {MASE: 0.558 $\rightarrow$ \textbf{0.520}};
  \node[casegaintext] at (1.905,0.33) {$\downarrow$\,6.70\%};
\end{scope}

\begin{scope}[shift={(12.15,0)}]
  \node[casetitle,text=caseviolet] at (0.20,4.43) {(d) Solar counters};
  \node[casekind,text=caseviolet] at (0.20,4.10) {Measurement structure};
  \node[casecue] at (0.20,3.78)
    {Daily readings accumulate,\\plateau, then reset.};
  \node[caseflow] at (1.905,2.775)
    {Forecast $\rightarrow$ Restore\\cumulative-reset shape};
  \node[casecode] at (0.20,2.20)
    {s = fit\_reset\_profile(y)\\
     level = daily\_level(q, s)\\
     q = level * s};
  \node[casescore] at (1.905,0.65) {MASE: 0.574 $\rightarrow$ \textbf{0.558}};
  \node[casegaintext] at (1.905,0.33) {$\downarrow$\,2.86\%};
\end{scope}
\end{tikzpicture}

%% file: sections/benchmark-experiments.tex
\paragraph{Data and Metrics.}
We evaluate all 98 tasks of \href{https://huggingface.co/spaces/Real-TSF/TIME-leaderboard}{\textbf{\timebench}} \citep{qiao2026itstimegenerationtime} and all 100 tasks of \href{https://huggingface.co/spaces/autogluon/fev-bench}{\textbf{\fevbench}} \citep{shchur2026fevbenchrealisticbenchmarktime}, covering \textbf{111,177} and \textbf{235,039 forecast instances}, respectively. Each instance is a series--window--target combination. Our default configuration pairs \href{https://huggingface.co/google/timesfm-3.0-pytorch}{TimesFM~3.0} with Claude~Code (Claude Fable~5). Following \timebench, we report normalized mean absolute scaled error (MASE) and continuous ranked probability score (CRPS), where lower is better. Following \fevbench, we report Skill Scores for MASE, scaled quantile loss (SQL), weighted quantile loss (WQL), and weighted absolute percentage error (WAPE), where higher is better. See App.~\ref{app:benchmarks} for metric definitions and protocol details.

\textbf{Baselines.}
We compare with published results from all 32 submissions on \timebench and 31 on \fevbench, covering foundation models, supervised forecasters, and statistical methods. Major baselines include TimesFM~3.0 (Google)~\citep{pmlr-v235-das24c}, Chronos-2 (Amazon)~\citep{ansari2025chronos2univariateuniversalforecasting}, Toto~2.0 (Datadog)~\citep{khwaja2026toto20timeseries}, QiYao-M (Huawei)~\citep{cheng2026qiyaommultimodaltimeseries}, and TiRex-2 (NXAI)~\citep{podest2026tirex2generalizingtirexmultivariate}.

\textbf{Main results. \method achieves the best scores on both benchmarks across all six reported metrics} (Fig.~\ref{fig:benchmarks}). Relative to direct inference with the same frozen TimesFM~3.0, it reduces MASE by 1.52\% and CRPS by 1.61\% on \timebench. On \fevbench, it raises MASE Skill by 2.79 percentage points and SQL Skill by 2.26 points. App.~\ref{app:benchmarks} gives the full comparisons, including WQL and WAPE Skill. These results show that improving the deployment pipeline can improve forecasting accuracy without updating the foundation model.

\textbf{Ablation. Feedback-guided revision improves deployment.}
We compare \method with Non-Iterative Search, which proposes the same number of candidate pipelines and selects among them without feedback-guided revision (Tab.~\ref{tab:feedback-ablation}). Non-iterative search underperforms direct inference on both benchmarks, whereas \method improves it. These results support revising pipelines based on forecast errors, rather than only proposing and selecting candidates.

%% file: sections/discovered-pipelines.tex
\section{Discovered Deployment Strategies}
\label{app:cases}

\begin{table*}[t]
\centering
\small
\begin{tabular}{llrrrr}
\hline
 & & \multicolumn{2}{c}{Test MASE $\downarrow$} & \multicolumn{2}{c}{Relative error reduction} \\
Task & Horizon & Direct & \method & MASE & CRPS \\
\hline
Supply chain & 30 days & 0.557750 & 0.520378 & 6.70\% & 6.76\% \\
Parking availability & 1 week & 0.678730 & 0.641658 & 5.46\% & 5.37\% \\
Ocean currents & 72 hours & 0.871531 & 0.835129 & 4.18\% & 0.64\% \\
Solar counters & 72 hours & 0.574087 & 0.557693 & 2.86\% & 2.66\% \\
Auto-production seasonal factor & 12 months & 0.756537 & 0.233980 & 69.07\% & 66.37\% \\
COVID reports & 30 days & 3.138536 & 2.890121 & 7.91\% & 15.62\% \\
Atmospheric variables & 24 hours & 0.476581 & 0.466005 & 2.22\% & 1.22\% \\
\hline
\end{tabular}
\caption{\textbf{Different tasks lead to different deployment programs.} Complete-task test scores for the selected examples. Relative reductions compare \method with direct TimesFM inference. The auto-production case uses an agent-written specialist for point forecasts and the foundation model for distributional information.}
\label{tab:case-results}
\end{table*}

These cases illustrate task-specific features, domain-informed representations, business rules, measurement structure, and auxiliary predictors. Tab.~\ref{tab:case-results} compares seven example tasks with direct inference using the same frozen TimesFM~3.0 backbone. Each test score covers all official windows and targets of its task. Historical adaptation and selection scores are reported separately; the selected examples do not represent the aggregate benchmark result.

\subsection{Supply Chain: Learning When Demand Is Zero}
\label{app:supply}

A holiday can break an otherwise regular weekly pattern. On July 4, 2024, in the first official forecast window of \texttt{Supply\_Chain\_Customer}, the first customer's observed order count is zero. The direct model's median forecast is $195.23$; \method's is zero.

The agent writes a date-based correction rather than a general low-demand threshold. It checks past observations on July 4, Thanksgiving, Christmas, and New Year's Day. For each target, the correction requires at least four past holiday observations, of which at least 80\% must be zero. When this condition holds, it sets all nine forecast quantiles to zero on the relevant holiday. Thus, a calendar hypothesis becomes a rule supported by that target's history.

The agent also tests corrections for the day after a holiday. Historical feedback leads it to restrict this adjustment to the day after July 4 rather than applying it to dates such as Black Friday. Further revisions align the normalized forecast shapes of orders, pieces, and revenue for the same customer. Tab.~\ref{tab:supply-revisions} shows all six revisions in this search. The main adaptation gain comes from the first holiday-aware correction; later edits refine it.

\begin{table}[t]
\centering
\small
\begin{tabular}{lr}
\hline
Revision & MASE $\downarrow$ \\
\hline
Starting pipeline & 0.522271 \\
1. Holiday and weekday corrections & 0.489970 \\
2. Post-holiday adjustment & 0.490036 \\
3. Align related forecast shapes & 0.489901 \\
4. Stronger post-holiday adjustment & 0.489933 \\
5. Restrict adjustment to July 5 & 0.489834 \\
6. Refine shape alignment & 0.489810 \\
\hline
\end{tabular}
\caption{Supply-chain revisions evaluated on the same adaptation windows. The starting pipeline already contains earlier adaptations; its score is not the direct-inference baseline.}
\label{tab:supply-revisions}
\end{table}

The task contains 36 targets and 12 windows of 30 days. On the four relevant holiday dates, all 144 target values are zero; direct inference predicts nonzero values, while \method predicts zero. Across all dates and targets, the complete pipeline reduces MASE by 6.70\%, including its other corrections.

\subsection{Parking: Revising How the Model Uses the Calendar}
\label{app:parking}

The task predicts available parking spaces every 15 minutes for the next week. Calendar information can describe more than a change in average availability: the daily pattern may also differ between weekdays and weekends. The agent explores both using a separate calendar predictor and supplying calendar features to TimesFM.

Tab.~\ref{tab:parking-revisions} shows all six evaluated revisions. The first calendar-predictor blend increases adaptation error. The agent then removes its contribution and builds known-future calendar inputs instead. The selected program contains 22 features: daily and weekly sine and cosine terms, weekend and Saturday indicators, and weekend--time-of-day interactions. Features cover both historical and future timestamps. The program includes a periodic feature only when the history spans at least 1.5 of its cycles.

\begin{table}[t]
\centering
\small
\begin{tabular}{lrr}
\hline
Revision & Adaptation & Selection \\
\hline
Direct TimesFM & 0.9421 & 0.6142 \\
1. Calendar-expert blend & 0.9578 & 0.6171 \\
2. Calendar inputs & 0.9208 & 0.6038 \\
3. History-length condition & 0.9208 & 0.6038 \\
4. Weekend--time features & 0.9268 & \textbf{0.6010} \\
5. Median and range edits & 0.9208 & 0.6038 \\
6. Fewer calendar inputs & \textbf{0.9152} & 0.6021 \\
\hline
\end{tabular}
\caption{Parking MASE on adaptation and separate selection windows; lower is better. Adaptation uses three origins and selection uses five, each with 32 car parks. Comparisons are within columns. Revision 4 is deployed, rather than the last revision or the adaptation winner.}
\label{tab:parking-revisions}
\end{table}

The chosen pipeline keeps the calendar predictor's output weight at zero: the improvement comes from a program that supplies new inputs to the frozen model. On the full test task of 354 car parks, MASE falls from $0.678730$ to $0.641658$ and CRPS from $0.046609$ to $0.044105$. The trace shows feedback changing the agent's approach, while separate historical selection determines which resulting program is deployed.

\begin{figure}[t]
  \input{figures/parking-features}
  \caption{\textbf{Calendar interactions in the generated program.} The core feature code, condensed for readability. Here, \texttt{tod} is the fraction of a day and \texttt{dow} ranges from Monday (0) to Sunday (6). These features cover both history and forecast dates.}
  \label{fig:parking-features}
\end{figure}

\subsection{Ocean Currents: Using Tidal Knowledge in Model Inputs}
\label{app:ocean}

In \texttt{current\_velocity/10T/long}, the agent uses known tidal periods to separate predictable oscillations from the remaining signal. Its program includes tidal constituents such as M2 (12.42 hours) and K1 (23.93 hours). It selects constituents whose frequencies can be distinguished over the available history and fits their amplitudes and phases for each target. Before applying the decomposition, a historical backtest checks whether the tidal fit improves on a constant-mean predictor.

The program subtracts the fitted tidal component from the history, supplies the residual to TimesFM, and adds the extrapolated component back to the point and quantile forecasts. Thus, domain knowledge changes the model's numerical input rather than only adding a textual description. The selected pipeline reduces MASE from $0.871531$ to $0.835129$ (4.18\%) and CRPS from $0.278207$ to $0.276433$ (0.64\%). These scores cover ten series, six variables, and five official windows, each forecasting 432 ten-minute steps (72 hours).

\subsection{Solar: Respecting a Daily-Reset Pattern}
\label{app:solar}

In \texttt{Australia\_Solar}, the agent identifies a pattern consistent with a daily-reset counter: values rise within the day, reach a plateau, and reset. This observed structure motivates a different correction from one designed for instantaneous power.

The program estimates a normalized daily profile $s(h)$ from the previous 28 valid days. For each forecast day and quantile, it fits a level $T$ to the model output and reconstructs the path as $T s(h)$. The correction preserves the historical daily shape and reset while using the model forecast to set the level.

The agent also explores daily-total experts and seasonal profiles. Separate historical selection favors the simpler shape projection. The selected pipeline reduces MASE by 2.86\% and CRPS by 2.66\% across three targets and 35 official windows, each covering 72 hours.

\subsection{Auto Production: Building a Calendar-Aware Specialist}
\label{app:auto}

The official task \texttt{Auto\_Production\_SF/M/short} contains the monthly seasonal-factor series \texttt{RSF\_AutoProduction}, not vehicle production counts. The agent builds on a calendar-aware predictor and explores holiday interactions, month-specific trends, and reversible seasonal transformations. It uses historical forecasts to choose a smoothing rate for each month, then shrinks these choices toward a common default.

The resulting specialist supplies the point forecast. TimesFM's quantile deviations and historical forecast errors inform the predictive distribution. This case illustrates the broader action space of deployment code: the agent can fit an auxiliary predictor, not only add input features.

Across all five 12-month test windows, MASE falls from $0.756537$ for direct inference to $0.233980$ for \method, a 69.07\% reduction. The pipeline used to start the final search already achieves $0.313005$; the six further revisions reduce its test MASE by 25.25\%. These comparisons distinguish the complete deployment gain from the additional gain of that search.

\subsection{COVID Reports: Revising an Overextended Trend}
\label{app:covid}

A plausible first idea need not be the best program. On \texttt{ECDC\_COVID}, the agent starts with a reversible logarithmic transformation, but adaptation MASE increases from $3.9173$ to $4.1816$. It then examines past forecast errors and finds overly strong upward extrapolation in the cases it inspects.

The next revision blends the model with a damped trend predictor, lowering adaptation MASE to $2.9624$. Further edits handle isolated reporting spikes and repeated values at the end of the history. The agent also tests shorter context and adjusts positive trend damping according to how long the recent growth has lasted. Tab.~\ref{tab:covid-revisions} gives the complete six-revision trace.

\begin{table}[t]
\centering
\small
\begin{tabular}{lrr}
\hline
Revision & Adaptation & Selection \\
\hline
Direct TimesFM & 3.9173 & 5.0403 \\
1. Log transformation & 4.1816 & 4.2079 \\
2. Damped trend blend & 2.9624 & 3.8822 \\
3. Repeated-value handling & 2.7920 & 4.1122 \\
4. Shorter context & 2.7604 & 4.1589 \\
5. Growth-duration damping & \textbf{2.7252} & 4.1114 \\
6. Milder damping and blend & 2.8398 & \textbf{3.7878} \\
\hline
\end{tabular}
\caption{COVID-report MASE on separate historical sets. Selection favors revision 6, which uses milder damping, over revision 5, which has the lowest adaptation error.}
\label{tab:covid-revisions}
\end{table}

The selected pipeline gives equal weight to the model and the auxiliary predictor in log space. Across nine countries and five 30-day test windows, it reduces task-level MASE by 7.91\% and CRPS by 15.62\%. These are aggregate gains across the task, not improvements in every country or window.

\subsection{Atmospheric Variables: Separating the Daily Cycle}
\label{app:weather}

\texttt{NE\_China\_Wind} forecasts four atmospheric variables, including wind components, geopotential height, and temperature. The agent fits daily harmonics in overlapping historical segments, allowing their amplitudes to change over time. It subtracts the estimated cycle, lets TimesFM forecast the remaining signal, and restores the future cycle afterward.

The transformation is reversible and fitted from the available history. The selected program reduces MASE by 2.22\% across all four variables and 30 official 24-hour windows.

%% file: figures/parking-features.tex
\begin{minipage}{\columnwidth}
\hrule
\smallskip
\small
\begin{verbatim}
weekend = np.where(dow >= 5, 1., -1.)
for k in (1, 2, 3):
    phase = 2 * np.pi * k * tod
    rows += [weekend * np.sin(phase),
             weekend * np.cos(phase)]
\end{verbatim}
\hrule
\end{minipage}

%% file: sections/benchmark-details.tex
\section{Benchmark Protocol and Full Results}
\label{app:benchmarks}
\label{app:benchmark-details}

\begin{table*}[t]
  \centering
  \input{sections/benchmark-tables}
  \caption{\textbf{Full public-result comparison.} All visible methods in the captured snapshot. Bold marks \method. \timebench values are normalized errors; \fevbench values are Skill Scores. \timebench includes the accepted entry; the \fevbench \method row is the submitted, not yet merged result. Rows are ordered by \timebench MASE and \fevbench SQL Skill.}
  \label{tab:full-benchmarks}
\end{table*}

\paragraph{Evaluation scope.}
\timebench provides 98 dataset--frequency--horizon tasks from 50 datasets, covering 111,177 forecast instances. \fevbench provides 100 tasks; our results cover all 1,308 specified forecast windows and 235,039 forecast instances. One instance is one series--window--target combination. We follow the released task definitions, including their target columns, sampling frequencies, horizons, forecast origins, and seasonal periods. All target variables and test windows contribute to each task's score. Available covariates retain their official roles: past-only covariates stop at the forecast origin, while known-future covariates may extend across the horizon.

\paragraph{Pipeline configuration.}
The default backbone is \href{https://huggingface.co/google/timesfm-3.0-pytorch}{TimesFM~3.0}. Its weights stay fixed throughout search and deployment. Our default code agent is Claude~Code (Claude Fable~5). The frozen candidate archive also includes programs developed with Claude Opus~5. Agents can use task and variable descriptions, timestamps, observed values, and available covariates. The editable stages are listed in App.~\ref{app:interface}; the historical selector is specified in App.~\ref{app:selector}. Predictions contain nine quantiles, at levels $0.1,0.2,\ldots,0.9$; the median supplies the point forecast.

\paragraph{Agent search budget.}
The default loop allows up to six iterations, with one candidate pipeline evaluated per iteration. Claude~Code uses high reasoning effort (\texttt{--effort high}), with a maximum of 256 agent turns and a 90-minute timeout per search session.

\paragraph{Chronological feedback.}
Historical adaptation windows provide code-search feedback, and a separate later block evaluates candidates. At deployment, only windows whose entire outcome has arrived can supply selection feedback. The forecast being evaluated cannot supply its own target values to the pipeline or selector. Fixed code may fit auxiliary predictors to the history available at the current origin. This is task-adapted deployment with a frozen backbone, rather than a claim of zero-shot operation for the whole system.

\paragraph{Point forecasting: MASE.}
Consider one target and forecast origin. Let $x_{1:L}$ be its observed history, $y_{1:H}$ its future outcomes, and $q_{\tau,h}$ its forecast at quantile level $\tau$ and lead $h$. The point forecast is $q_{0.5,h}$. With seasonal period $s$, the historical scale and mean absolute scaled error are
\begin{align}
 a_s &= \frac{1}{L-s}\sum_{\ell=s+1}^{L}|x_\ell-x_{\ell-s}|,\\
 \operatorname{MASE} &= \frac{1}{H a_s}\sum_{h=1}^{H}|y_h-q_{0.5,h}|.
\end{align}
Lower is better. The seasonal scale makes errors comparable across targets with different units.

\paragraph{Probabilistic forecasting: SQL and CRPS.}
Let $\mathcal{Q}=\{0.1,0.2,\ldots,0.9\}$. Define the doubled quantile loss as
\begin{equation}
 L_\tau(y,q)=2\bigl(\tau-\mathbf{1}[y<q]\bigr)(y-q).
\end{equation}
\fevbench's scaled quantile loss divides by the historical seasonal scale:
\begin{equation}
 \operatorname{SQL}=\frac{\sum_{\tau\in\mathcal{Q}}\sum_{h=1}^{H}L_\tau(y_h,q_{\tau,h})}{|\mathcal{Q}| H a_s}.
\end{equation}
\timebench reports a quantile approximation to the continuous ranked probability score (CRPS), normalized by the target magnitude. Its implementation uses mean weighted quantile loss:
\begin{equation}
 \operatorname{CRPS}_{\mathrm{TIME}}=
 \frac{\sum_{\tau\in\mathcal{Q}}\sum_{h=1}^{H}L_\tau(y_h,q_{\tau,h})}{|\mathcal{Q}|\sum_{h=1}^{H}|y_h|}.
\end{equation}
Both are lower-is-better. The distinction is the denominator: SQL uses past seasonal differences; \timebench's CRPS uses the magnitude of the outcomes being scored. We use the official evaluators to aggregate over series, targets, and windows, retaining their handling of missing values and undefined scales.

\paragraph{\timebench aggregation.}
Let $E_{i,m}$ be the official task-level error for metric $m$ and $B_{i,m}$ the corresponding Seasonal Naive error. Define $r_{i,m}=E_{i,m}/B_{i,m}$. \timebench reports
\begin{equation}
 R_m=\exp\left(\frac{1}{N}\sum_{i=1}^{N}\log r_{i,m}\right),\quad N=98.
\end{equation}
Lower is better and Seasonal Naive scores 1. The public \timebench table displays three decimal places. Tab.~\ref{tab:full-benchmarks} preserves that precision, while Tab.~\ref{tab:paired-backbone} uses our saved full-precision aggregates.

\paragraph{\fevbench aggregation.}
Using the same task-relative error notation, \fevbench reports the Skill Score $S_m$:
\begin{equation}
 S_m=100\left[1-\exp\left(\frac{1}{N}\sum_{i=1}^{N}\log\widetilde{r}_{i,m}\right)\right],
 \label{eq:fev-skill}
\end{equation}
where $N=100$ and $\widetilde{r}_{i,m}=\min(100,\max(0.01,r_{i,m}))$. Higher is better; 0 matches Seasonal Naive. We retain the official missing-result and pretraining-overlap handling: missing results use Seasonal Naive and flagged pretraining overlap uses Chronos-Bolt. Models marked hidden in the public metadata are omitted, matching the leaderboard's display policy.

\paragraph{Baseline sources.}
The public comparison includes TimesFM \citep{pmlr-v235-das24c}, Chronos-2 \citep{ansari2025chronos2univariateuniversalforecasting}, Toto~2.0 \citep{khwaja2026toto20timeseries}, QiYao-M \citep{cheng2026qiyaommultimodaltimeseries}, and TiRex-2 \citep{podest2026tirex2generalizingtirexmultivariate}. These citations identify the model families; the exact released versions are named in Tab.~\ref{tab:full-benchmarks}. API and release-only entries, such as \href{https://huggingface.co/yhatlabs/ChakraTS}{ChakraTS}, follow the official model cards and benchmark release records \citep{qiao2026itstimegenerationtime,shchur2026fevbenchrealisticbenchmarktime}. We use their published scores, rather than rerunning or retuning their configurations.

\paragraph{Comparison scope.}
Public baseline scores use the October 10, 2026 snapshot of both benchmarks. They include pretrained forecasters and, where available, task-trained and statistical methods. The comparison uses their published configurations, not a common compute budget. \timebench has accepted our entry; the \fevbench table adds our pending submission to the public baseline files. Tab.~\ref{tab:paired-backbone} separately compares \method with direct inference in our environment.

\begin{table}[t]
  \centering
  \small
  \begin{tabular}{lrrr}
    \hline
    Metric & Direct & \method & Gain \\
    \hline
    \multicolumn{4}{l}{\timebench: normalized error $\downarrow$} \\
    MASE & 0.639759 & 0.630011 & 1.52\% \\
    CRPS & 0.536328 & 0.527713 & 1.61\% \\
    \multicolumn{4}{l}{\fevbench: Skill Score (\%) $\uparrow$} \\
    MASE & 37.786 & 40.579 & 4.49\% \\
    SQL & 48.984 & 51.243 & 4.43\% \\
    \hline
  \end{tabular}
  \caption{\textbf{Paired comparison with direct inference.} The reference uses our environment and the same backbone. Gain denotes relative error reduction. For \fevbench it is computed from $100-\text{Skill}$. The public FEV TimesFM row differs slightly from this local reference.}
  \label{tab:paired-backbone}
\end{table}

\paragraph{Generalization across agents and TSFMs.}
We report additional \timebench configurations in Tab.~\ref{tab:generalization}. The agent comparison uses TimesFM~3.0 with Claude~Code (Claude Fable~5) or Antigravity (Gemini~3.8~Flash). The backbone comparison uses Claude~Code (Claude Fable~5) to evaluate \method with Chronos-2 and LPTM against direct inference with the same backbone. Scores use the same normalized MASE and CRPS metrics as the main comparison and are reported to three decimal places. On all 100 \fevbench tasks, \method with LPTM reaches SQL, MASE, WQL, and WAPE Skill Scores of 46.33\%, 34.63\%, 50.96\%, and 38.58\%, compared with 43.20\%, 30.70\%, 47.60\%, and 34.18\% for direct LPTM inference. These experiments evaluate the framework with different components.

\begin{table}[t]
  \centering
  \small
  \begin{tabular}{lll}
    \toprule
    Configuration & MASE $\downarrow$ & CRPS $\downarrow$ \\
    \midrule
    \multicolumn{3}{l}{TimesFM~3.0 backbone} \\
    Direct inference & 0.640 & 0.536 \\
    \method: Claude~Code & \textbf{0.630} & \textbf{0.528} \\
    \method: Antigravity & 0.637 & 0.532 \\
    \midrule
    \multicolumn{3}{l}{Chronos-2 backbone} \\
    Direct inference & 0.662 & 0.556 \\
    \method: Claude~Code & \textbf{0.647} & \textbf{0.545} \\
    \midrule
    \multicolumn{3}{l}{LPTM backbone} \\
    Direct inference & 0.669 & 0.572 \\
    \method: Claude~Code & \textbf{0.641} & \textbf{0.549} \\
    \bottomrule
  \end{tabular}
  \caption{\textbf{Deployment gains across coding agents and TSFMs.} Normalized errors on all 98 \timebench tasks. Claude~Code uses Claude Fable~5; Antigravity uses Gemini~3.8~Flash. Bold marks the best scores within each backbone group.}
  \label{tab:generalization}
\end{table}

\subsection{Additional FEV-bench Metrics}
\label{app:fev-additional-metrics}

Tab.~\ref{tab:fev-all-metrics} gives the complete comparison across the four \fevbench metrics. \method ranks first in each, including WQL Skill of 55.759\% and WAPE Skill of 44.788\%.

\begin{table*}[t]
  \centering
  \small
  \input{sections/fev-all-metric-table}
  \caption{\textbf{Full \fevbench comparison across four metrics.} Skill Scores (\%; higher is better) cover all 100 tasks. We include all 31 publicly visible baselines and our submitted result. Rows are ordered by SQL Skill; bold marks the best score in each column.}
  \label{tab:fev-all-metrics}
\end{table*}

\paragraph{Metric definitions.}
For one target variable in a forecast window, let $i$ index series and $h$ index future time steps. Using the quantile loss $L_\tau$ and levels $\mathcal{Q}$ defined above, weighted quantile loss is
\begin{equation}
 \operatorname{WQL} =
 \frac{\sum_{\tau\in\mathcal{Q}}\sum_{i,h}L_\tau(y_{i,h},q_{\tau,i,h})}
 {|\mathcal{Q}|\sum_{i,h}|y_{i,h}|}.
\end{equation}
Weighted absolute percentage error uses the point forecast $\widehat{y}_{i,h}$:
\begin{equation}
 \operatorname{WAPE} =
 \frac{\sum_{i,h}|y_{i,h}-\widehat{y}_{i,h}|}
 {\sum_{i,h}|y_{i,h}|}.
\end{equation}
The official evaluator averages scores over target variables and forecast windows. Lower raw WQL and WAPE are better; their Skill Scores follow Eq.~\ref{eq:fev-skill}, with higher values indicating better performance relative to Seasonal Naive. We apply the same missing-result, pretraining-overlap, and visibility rules as for SQL and MASE.

%% file: sections/benchmark-tables.tex
\begin{minipage}[t]{0.48\textwidth}
\centering\small
\textbf{\timebench: normalized errors}\par\smallskip
\begin{tabular}{lrr}
\hline
Method & MASE $\downarrow$ & CRPS $\downarrow$ \\
\hline
\textbf{\method} & \textbf{0.630} & \textbf{0.528} \\
QiYao-M & 0.638 & 0.536 \\
ChakraTS & 0.639 & 0.537 \\
TimesFM 3.0 & 0.640 & 0.536 \\
Toto 2.0 (2.5B) & 0.642 & 0.539 \\
Toto 2.0 (313M) & 0.644 & 0.542 \\
Toto 2.0 (1B) & 0.645 & 0.545 \\
PatchTST-FM-Extended & 0.662 & 0.557 \\
Chronos-2 & 0.662 & 0.556 \\
T0-beta & 0.665 & 0.555 \\
Tabby & 0.668 & 0.553 \\
TimesFM 2.5 & 0.669 & 0.567 \\
OmniScient & 0.669 & 0.562 \\
Toto 2.0 (22M) & 0.670 & 0.562 \\
PatchTST-FM-R1 & 0.671 & 0.564 \\
Tabby-Pretrain & 0.678 & 0.566 \\
TiRex & 0.683 & 0.573 \\
Toto 2.0 (4M) & 0.691 & 0.582 \\
TS-ICL & 0.694 & 0.581 \\
Toto & 0.698 & 0.584 \\
Eidos & 0.700 & 0.587 \\
Timer-S1 & 0.702 & 0.589 \\
Moirai 2.0 & 0.703 & 0.588 \\
TimesFM 2.0 & 0.718 & 0.620 \\
Chronos-Bolt & 0.733 & 0.620 \\
Kairos & 0.739 & 0.644 \\
sundial\_base & 0.758 & 0.663 \\
Fracast-0 & 0.768 & 0.649 \\
litespecformer & 0.787 & 0.653 \\
TimesFM 1.0 & 0.788 & 0.689 \\
visiontspp\_base & 0.805 & 0.671 \\
Moirai\_base & 0.826 & 0.666 \\
seasonal\_naive & 1.000 & 1.000 \\
\hline
\end{tabular}
\end{minipage}
\hfill
\begin{minipage}[t]{0.48\textwidth}
\centering\small
\textbf{\fevbench: Skill Score (\%)}\par\smallskip
\begin{tabular}{lrr}
\hline
Method & MASE $\uparrow$ & SQL $\uparrow$ \\
\hline
\textbf{\method} & \textbf{40.579} & \textbf{51.243} \\
TimesFM 3.0 & 37.816 & 48.994 \\
ChakraTS & 36.930 & 48.143 \\
Chronos-2 & 35.496 & 47.281 \\
Lingjiang2.0 & 35.377 & 47.020 \\
T0-beta & 34.863 & 46.653 \\
TiRex-2 & 33.743 & 45.499 \\
Toto 2.0 (2.5B) & 32.539 & 44.421 \\
TabPFN-TS-3 & 30.562 & 43.150 \\
TS-ICL & 31.113 & 43.115 \\
Toto 2.0 (22M) & 30.609 & 42.907 \\
TiRex & 30.012 & 42.576 \\
TimesFM 2.5 & 30.198 & 42.201 \\
TabPFN-TS & 29.834 & 41.487 \\
CITRAS-FM & 29.329 & 41.178 \\
Toto 1.0 & 28.233 & 40.755 \\
FlowState & 27.830 & 39.911 \\
Moirai 2.0 & 27.225 & 39.314 \\
Chronos-Bolt & 26.515 & 38.893 \\
Sundial-Base & 24.746 & 33.423 \\
TFT & 20.479 & 32.179 \\
PatchTST & 18.365 & 30.089 \\
DeepAR & 17.521 & 29.084 \\
CatBoost & 23.692 & 22.998 \\
AutoARIMA & 11.626 & 21.304 \\
LightGBM & 21.692 & 20.979 \\
Stat. Ensemble & 16.443 & 20.655 \\
AutoTheta & 10.988 & 5.457 \\
Seasonal Naive & 0.000 & 0.000 \\
AutoETS & 2.259 & -26.819 \\
Naive & -16.674 & -45.399 \\
Drift & -18.139 & -45.776 \\
\hline
\end{tabular}
\end{minipage}

%% file: sections/fev-all-metric-table.tex
\begin{tabular*}{\textwidth}{@{\extracolsep{\fill}}lllll@{}}
\toprule
Method & SQL Skill $\uparrow$ & MASE Skill $\uparrow$ & WQL Skill $\uparrow$ & WAPE Skill $\uparrow$ \\
\midrule
\method & \textbf{51.243} & \textbf{40.579} & \textbf{55.759} & \textbf{44.788} \\
TimesFM 3.0 & 48.994 & 37.816 & 53.278 & 41.656 \\
ChakraTS & 48.143 & 36.930 & 52.440 & 40.576 \\
Chronos-2 & 47.281 & 35.496 & 51.520 & 39.410 \\
Lingjiang2.0 & 47.020 & 35.377 & 51.457 & 39.229 \\
T0-beta & 46.653 & 34.863 & 50.827 & 38.318 \\
TiRex-2 & 45.499 & 33.743 & 49.542 & 37.014 \\
Toto 2.0 (2.5B) & 44.421 & 32.539 & 49.217 & 36.655 \\
TabPFN-TS-3 & 43.150 & 30.562 & 48.824 & 35.887 \\
TS-ICL & 43.115 & 31.113 & 47.233 & 34.312 \\
Toto 2.0 (22M) & 42.907 & 30.609 & 47.668 & 34.553 \\
TiRex & 42.576 & 30.012 & 46.666 & 33.560 \\
TimesFM 2.5 & 42.201 & 30.198 & 46.726 & 33.719 \\
TabPFN-TS & 41.487 & 29.834 & 46.646 & 34.853 \\
CITRAS-FM & 41.178 & 29.329 & 45.997 & 33.229 \\
Toto 1.0 & 40.755 & 28.233 & 45.021 & 31.521 \\
FlowState & 39.911 & 27.830 & 46.006 & 33.122 \\
Moirai 2.0 & 39.314 & 27.225 & 43.852 & 30.624 \\
Chronos-Bolt & 38.893 & 26.515 & 43.187 & 29.770 \\
Sundial-Base & 33.423 & 24.746 & 37.437 & 27.263 \\
TFT & 32.179 & 20.479 & 34.969 & 22.701 \\
PatchTST & 30.089 & 18.365 & 34.333 & 21.454 \\
DeepAR & 29.084 & 17.521 & 32.580 & 20.002 \\
CatBoost & 22.998 & 23.692 & 26.618 & 25.514 \\
AutoARIMA & 21.304 & 11.626 & 24.478 & 13.599 \\
LightGBM & 20.979 & 21.692 & 24.360 & 23.221 \\
Stat. Ensemble & 20.655 & 16.443 & 22.373 & 19.454 \\
AutoTheta & 5.457 & 10.988 & 7.846 & 13.795 \\
Seasonal Naive & 0.000 & 0.000 & 0.000 & 0.000 \\
AutoETS & -26.819 & 2.259 & -27.027 & 4.329 \\
Naive & -45.399 & -16.674 & -39.121 & -6.111 \\
Drift & -45.776 & -18.139 & -40.059 & -8.605 \\
\bottomrule
\end{tabular*}